\documentclass[conference]{IEEEtran}
\IEEEoverridecommandlockouts
\usepackage{cite}
\usepackage{amsmath,amssymb,amsfonts}
\usepackage{algorithmic}
\usepackage{graphicx}
\usepackage{textcomp}
\usepackage{xcolor}

\usepackage[colorinlistoftodos]{todonotes}
\usepackage{url}
\usepackage{algorithm}
\usepackage{algorithmic}

\usepackage{booktabs}

\usepackage{subfig}

\def\BibTeX{{\rm B\kern-.05em{\sc i\kern-.025em b}\kern-.08em
    T\kern-.1667em\lower.7ex\hbox{E}\kern-.125emX}}
\begin{document}

\title{Transfer learning for time series classification}


\author{\IEEEauthorblockN{Hassan Ismail Fawaz,
Germain Forestier,
Jonathan Weber,
Lhassane Idoumghar and 
Pierre-Alain Muller\smallskip }
\IEEEauthorblockA{IRIMAS,
Universit\'e de Haute-Alsace, Mulhouse, France \smallskip}
\IEEEauthorblockN{\textit{Email}: \{first-name.last-name@uha.fr\}}
}

\newtheorem{definition}{Definition}


\maketitle

\begin{abstract}

Transfer learning for deep neural networks is the process of first training a base network on a source dataset, and then transferring the learned features (the network's weights) to a second network to be trained on a target dataset. 
This idea has been shown to improve deep neural network's generalization capabilities in many computer vision tasks such as image recognition and object localization. 
Apart from these applications, deep Convolutional Neural Networks (CNNs) have also recently gained popularity in the Time Series Classification (TSC) community. 
However, unlike for image recognition problems, transfer learning techniques have not yet been investigated thoroughly for the TSC task. 
This is surprising as the accuracy of deep learning models for TSC could potentially be improved if the model is fine-tuned from a pre-trained neural network instead of training it from scratch. 
In this paper, we fill this gap by investigating how to transfer deep CNNs for the TSC task. 
To evaluate the potential of transfer learning, we performed extensive experiments using the UCR archive which is  the largest publicly available TSC benchmark containing 85 datasets. 
For each dataset in the archive, we pre-trained a model and then fine-tuned it on the other datasets resulting in 7140 different deep neural networks. 
These experiments revealed that transfer learning can improve or degrade the model’s predictions depending on the dataset used for transfer. 
Therefore, in an effort to predict the best source dataset for a given target dataset, we propose a new method relying on Dynamic Time Warping to measure inter-datasets similarities.
We describe how our method can guide the transfer to choose the best source dataset leading to an improvement in accuracy on 71 out of 85 datasets.
\end{abstract}

\begin{IEEEkeywords}
Transfer learning, time series classification, deep learning, Dynamic Time Warping
\end{IEEEkeywords}

\section{Introduction}

\begin{figure}[htbp]
\centerline{\includegraphics[width=0.95\linewidth]{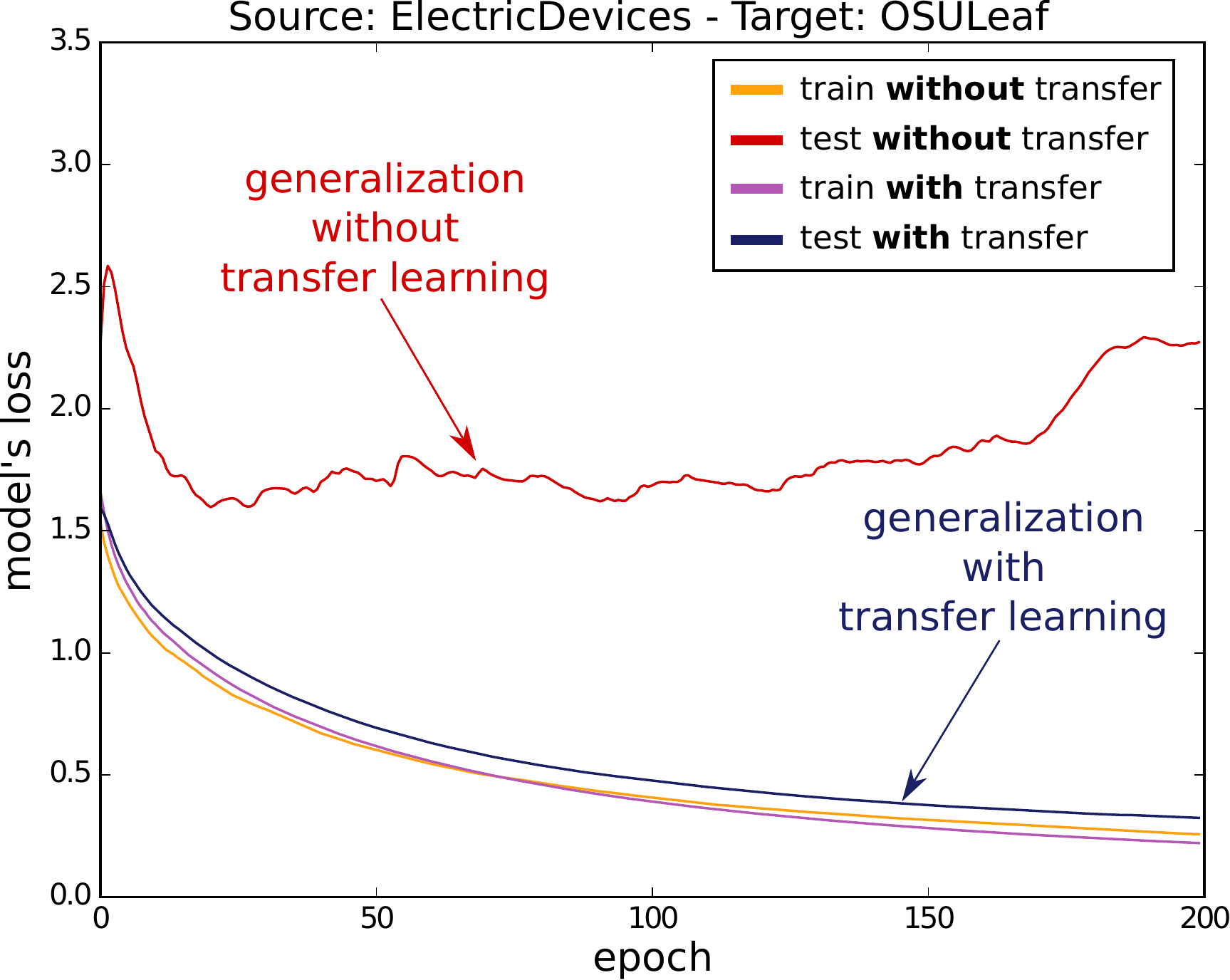}}
\caption{Evolution of model's loss (train and test) with and without the transfer learning method using ElectricDevices as source and OSULeaf as target datasets. 
(Best viewed in color).}
\label{fig-elect-shapelets}
\end{figure}

Convolutional Neural Networks (CNNs) have recently been shown to significantly outperform the nearest neighbor approach coupled with the Dynamic Time Warping (DTW) algorithm (1NN-DTW) on the UCR archive benchmark~\cite{ucrarchive} for the Time Series Classification (TSC) problem~\cite{wang2017time}. 
CNNs were not only able to beat the 1NN-DTW baseline, but they also reached results that are not significantly different than COTE~\cite{bagnal2015time} - which is an ensemble of 37 classifiers.
However, despite the high performance of these CNNs, deep learning models are still prone to overfitting. One example where these neural networks fail to generalize is when the training set of the time series dataset is very small.
For example, while the accuracy of the Fully Convolutional Neural Networks (FCN)~\cite{wang2017time} is 30\% on the DiatomSizeReduction dataset (whose training set is the smallest in the UCR archive~\cite{ucrarchive}), the 1NN-DTW classifier reaches 96\% on the same dataset with the same train-test split~\cite{ucrarchive}.
We attribute this huge difference in accuracy to the overfitting phenomena, which is still an open area of research in the deep learning community~\cite{zhang2016understanding}.  
This problem is known to be mitigated using several techniques such as regularization, data augmentation or simply collecting more data~\cite{zhang2016understanding,IsmailFawaz2018data}.
Another well-know technique is transfer learning~\cite{yosinski2014transferable}, where a model trained on a source task is then fine-tuned on a target dataset. 
For example in \figurename~\ref{fig-elect-shapelets}, we trained a model on the ElectricDevices dataset~\cite{ucrarchive} and then fine-tuned this same model on the OSULeaf dataset~\cite{ucrarchive}, which significantly improved the network's generalization capability.

Transfer learning is currently used in almost every deep learning model when the target dataset does not contain enough labeled data~\cite{yosinski2014transferable}.
Despite its recent success in computer vision~\cite{csurka2017domain}, transfer learning has been rarely applied to deep learning models for time series data.
One of the reasons for this absence is probably the lack of one big general purpose dataset similar to ImageNet~\cite{russakovsky2015imagenet} or OpenImages \cite{openimages} but for time series.
Furthermore, it is only recently that deep learning was proven to work well for TSC~\cite{cui2016multi} and there is still much to be explored in building deep neural networks for mining time series data~\cite{gamboa2017deep}.

Since transferring deep learning models, between the UCR archive datasets~\cite{ucrarchive} (the largest benchmark currently available), have not been thoroughly studied, we decided to investigate this area of research with the ultimate goal to determine in advance which dataset transfers could benefit the CNNs and improve their TSC accuracy. 

The intuition behind the transfer learning approach for time series data is also partially inspired by the observation of Cui~et~al.~\cite{cui2016multi}, where the authors showed that shapelets~\cite{ye2009time} (or subsequences) learned by the learning shapelets approach~\cite{grabocka2014learning} are related to the filters (or kernels) learned by the CNNs.
We hypothesize that these learned subsequences might not be specific to one dataset and could occur in other unseen datasets with un/related classification tasks.
Another observation for why transfer learning should work for time series data is its recent success in computer vision tasks~\cite{csurka2017domain}. 
Indeed, since time series data contain one temporal dimension (time) compared to two dimensions for images (width and height), it is only natural to think that if filters can successfully be transferred on images~\cite{yosinski2014transferable}, they should also be transferable across time series datasets. 

\begin{figure}
\centering
    \includegraphics[width=\linewidth]{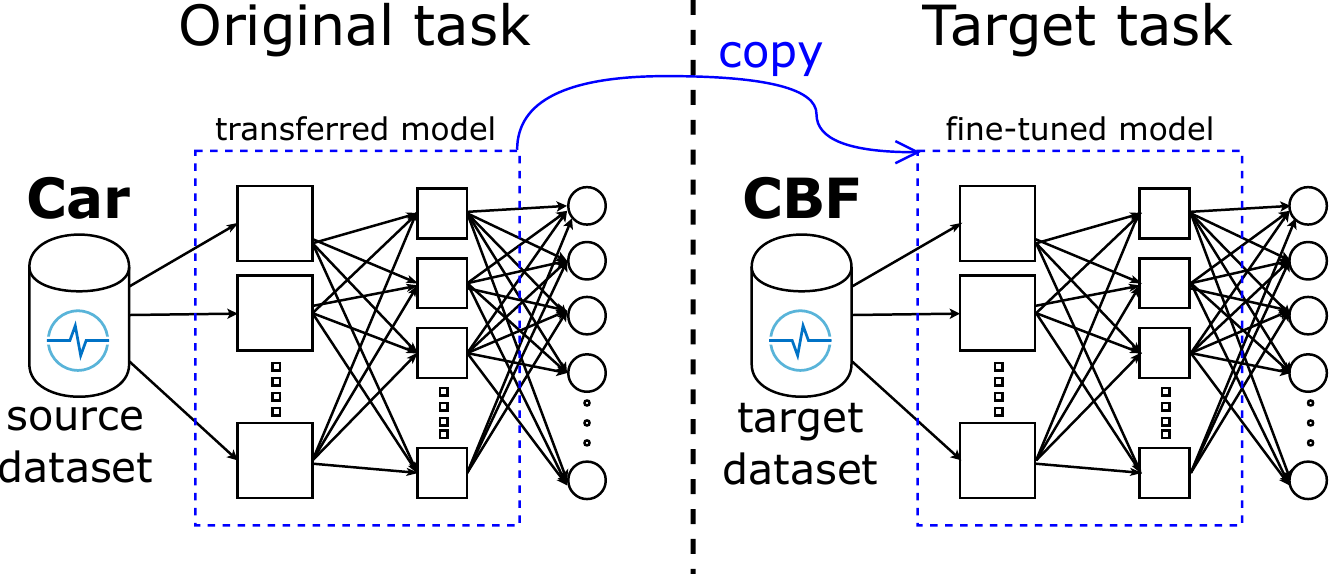}
    \caption{General deep learning training process with transfer learning for time series classification.
    In this example, a model is first pre-trained on Car (source dataset) and then the corresponding weights are fine-tuned on CBF (target dataset).}
    \label{fig-transfer_archi}
\end{figure}

To evaluate the potential of transfer learning for TSC, we performed experiments where each pair of datasets in the UCR archive was tested twice: we pre-trained a model for each dataset, then transferred and fine-tuned it on all the other datasets (a total of more than $7140$ trained models).
\figurename~\ref{fig-transfer_archi} illustrates the architecture of our proposed framework of transfer learning for TSC on two datasets. 
The obtained results show that time series do exhibit some low level features that could be used in a transfer learning approach. 
They also show that using transfer learning reduces the training time by reducing the number of epochs needed for the network to converge on the train set. 

Motivated by the consensus that transferring models between similar datasets improves the classifier's accuracy~\cite{weiss2016a}, we used the DTW algorithm as an inter-datasets similarity measure in order to quantify the relationship between the source and target datasets in our transfer learning framework.
Our experiments show that DTW can be used to predict the best source dataset for a given target dataset. 
Our method can thus identify which datasets should be considered for transfer learning given a new TSC problem.

The rest of the paper is structured as follows, in Section~\ref{sec-background} we review the existing work on deep and transfer learning for time series analysis.
We then detail our proposed framework in Section~\ref{sec-method}. 
In Section~\ref{sec-experiments}, we present the setups for our experimentations followed by the corresponding results and discussions in Section~\ref{sec-results}. 
Finally, in Section~\ref{sec-conclusion}, we conclude the work presented in this paper while proposing directions for future research.  

\section{Background and related work}\label{sec-background}

In this section, we start by giving a formal definition for the TSC problem. 
We then present some recent successful applications of deep neural networks for the TSC task.
Finally, we give a summary of transfer learning and some of its applications for time series data mining problems.

\subsection{Time Series Classification}
\begin{definition}
A time series $X=[x_1,x_2, ... x_T]$ is an ordered set of real values. 
The length of $X$ is equal to the number of real values $T$. 
\end{definition}

\begin{definition}
A dataset $D=\{(X_1,Y_1),\dots,(X_N,Y_N)\}$ is a collection of pairs $(X_i,Y_i)$ where $X_i$ is a time series with $Y_i$ as its corresponding label (or class) vector.  
\end{definition}

The task of TSC consists in training a classifier on a dataset $D$ in order to map from the space of possible inputs $X_i$ to a probability distribution over the class variable values $Y_i$.

\subsection{Deep learning for Time Series Classification}

Since AlexNet~\cite{krizhevsky2012imagenet} won the ImageNet competition in 2012, deep learning has seen a lot of successful applications in many different domains~\cite{lecun2015deep} such as reaching human level performance in image recognition problems~\cite{szegedy2015going} as well as different natural language processing tasks~\cite{sutskever2014sequence,bahdanau2014neural}. 
Motivated by this success of deep neural networks in many different domains, deep learning has been recently applied for the TSC problem~\cite{geng2018cost,leguennec2016data}.

In~\cite{wang2017time}, a three layers Fully Convolutional Neural Network has been designed for the TSC problem and validated on the UCR archive~\cite{ucrarchive}. 
The FCN architecture contains convolutional layers without any sub-sampling layer, followed by a global average pooling layer before a traditional softmax classifier as the final layer.
The network's architecture is presented in \figurename~\ref{fig-fcn_archi} and explained in details in Section~\ref{sec-method}.
Given that this network is currently the state of the art deep learning model for the TSC problem, we chose it as our main network for exploring transfer learning. 

Other deep learning models have been also validated on the UCR archive~\cite{ucrarchive}. 
In~\cite{leguennec2016data}, a deep CNN was designed and trained from scratch to classify time series.  
In order to avoid the overfitting problem, the authors proposed different data augmentation techniques that warped and split the time series. 
In~\cite{geng2018cost}, the authors took the FCN model and modified the cost function in order to take into account the imbalanced classification of time series. 

Outside the UCR archive~\cite{ucrarchive}, deep learning has reached state of the art performance on several datasets in different domains~\cite{langkvist2014a}. 
For spatio-temporal series forecasting problems, such as meteorology and oceanography, deep	 neural networks were proposed in~\cite{ziat2017spatio}.
For human activity recognition from wearable sensors, deep learning is replacing the feature engineering approaches~\cite{nweke2018deep} where features are no longer hand-designed but rather learned by deep learning models trained through back-propagation.
One other type of time series data is present in Electronic Health Records, where a recent generative adversarial network with a CNN~\cite{che2017boosting} was trained for risk prediction based on patients historical medical records.    

In short, deep learning is being applied to time series data with very successful results in several different domains.
In fact, the convolutional neural network's ability to learn temporal invariant features is one of main the reasons behind its recent success, as well as the availability of big data across different domains~\cite{IsmailFawaz2018deep}.

Given the nature of time series data in many real-life applications, a question arises: Could the knowledge discovered in a certain dataset, be leveraged in order to boost the performance of deep neural networks on another completely unrelated time series dataset ?

\subsection{Transfer learning for Time Series Classification}

Before getting into the details of the recent applications for transfer learning, we give a formal definition of the latter~\cite{weiss2016a}. 
\begin{definition}
Transfer learning for deep neural networks, is the process of first training a base network on a source dataset and task, and then transfer the learned features (the network's weights) to a second network to be trained on a target dataset and task. 
\end{definition}
Throughout this paper, we will refer to \emph{source} dataset as the dataset we are transferring the pre-trained model \emph{from}, and to \emph{target} dataset as the dataset we are transferring the pre-trained model \emph{to}.

Now that we have established the necessary definitions, we will dive into the recent applications of transfer learning for time series data mining tasks. 
In fact, transfer learning is sometimes confused with the domain adaptation approach~\cite{pan2010a,long2015learning}. 
The main difference with the latter method is that the model is jointly trained on the source and target datasets~\cite{weiss2016a}.
The goal of using the target instances during training, is to minimize the discrepancy between the source's and target's instances.
In~\cite{ariefang2017da}, a domain adaptation approach was proposed to predict human indoor occupancy based on the carbon dioxide concentration in the room.
In~\cite{kasteren08recognizing}, hidden Markov models' generative capabilities were used in a domain adaptation approach to recognize human activities based on a sensor network.   

For time series anomaly detection, a transfer learning approach was used to determine which time series should be transferred from the source to the target dataset to be used with a 1-NN DTW classifier~\cite{vercruyssen2017transfer}.
Similarly, in~\cite{spiegel2016} the authors developed a method to transfer specific training examples from the source dataset to the target dataset and hence compute the dissimilarity matrix using the new training set. 
As for time series forecasting, a transfer learning approach for an auto-encoder was employed to predict the wind-speed in a farm~\cite{hu2016transfer}. 
The authors proposed first to train a model on the historical wind-speed data of an old farm and fine-tune it using the data of a new farm. 
In~\cite{banerjee2017a} restricted Boltzmann machines were first pre-trained for acoustic phoneme recognition and then fine-tuned for post-traumatic stress disorder diagnosis.    

Perhaps the recent work in~\cite{serra2018towards} is the closest to ours in terms of using transfer learning to improve the accuracy of deep neural networks for TSC. 
In this work, the authors designed a CNN with an attention mechanism to encode the time series in a supervised manner. 
Before fine-tuning a model on a target dataset, the model is first jointly pre-trained on several source datasets with themes~\cite{bagnall2017the} that are different from the target dataset's theme which limits the choice of the source dataset to only one. 
Additionally, unlike~\cite{serra2018towards}, we take a pre-designed deep learning model without modifying it nor adding regularizers.   
This enabled us to solely attribute the improvement in accuracy to the transfer learning feature, which we describe in details in the following section.

\section{Method}\label{sec-method}

In this section, we present our proposed method of transfer learning for TSC. 
We first introduce the adopted neural network architecture from~\cite{wang2017time}. 
We then thoroughly explain how we adapted the network for the transfer learning process. 
Finally, we present our DTW based method that enabled us to compute the inter-datasets similarities, which we later use to guide the transfer learning process.  

\begin{figure*}[htbp]
\centerline{\includegraphics[width=0.88\linewidth]{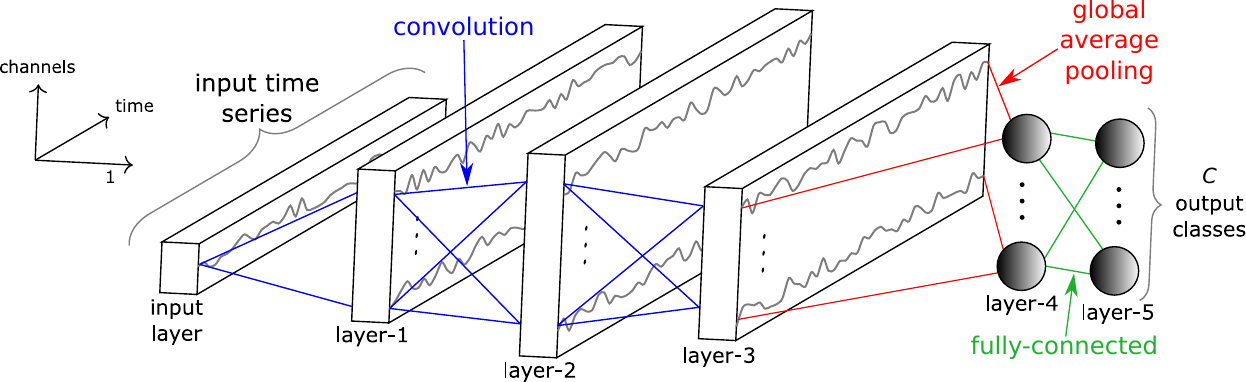}}
\caption{Architecture of the one dimensional Fully Convolutional Neural Network adopted for our transfer learning process.
The input to the network is a time series of variable length. 
The output layer is a softmax fully-connected classifier with $C$ neurons equal to the number of classes in the dataset. (Best viewed in color).
}
\label{fig-fcn_archi}
\end{figure*}

\subsection{Architecture}
The network architecture which we have selected for the transfer learning approach, is a one dimensional Fully Convolutional Neural Network~\cite{wang2017time} (FCN). 
Note that our transfer learning method is independent of the chosen network architecture, and that we have chosen this latter architecture for its robustness as it has already achieved state of the art results on 44 datasets from the UCR archive~\cite{ucrarchive}. 

The input of the network is a time series of variable length.
The network's output is a probability distribution over the $C$ possible classes in the dataset.  
The first, second and third layers are convolutional layers with the Rectified Linear Unit (ReLU) as activation function.
Each convolutional layer is followed by a batch normalization operation~\cite{ioffe2015batch}.
More precisely, the first convolutional layer is composed of 128 filters of length 8.
The second convolution is composed of 256 filters of length 5. 
The last convolutional layer contains 128 filters of length 3 and the three convolutions have a stride equal to 1. 

Each convolutional layer takes as input a time series and perform some non-linearities to transform it into a multivariate time series whose dimensions are inferred from the number of filters in each layer. 
The fourth layer is composed of a global average pooling operation which takes the input of the third convolution and averages each time series over the time axis. 
This averaging operation reduces drastically the number of parameters in a deep model while enabling the use of a class activation map~\cite{zhou2016learning} which allows an interpretation of the learned features~\cite{wang2017time}. 
The output of layer four is then fed to a softmax classification layer whose number of neurons is equal to the number of classes $C$ in the dataset. 
The rest of the network's hyperparameters are shown in Table~\ref{tab-hyper}. 

Similarly to the hyperparameters, the architecture depicted in \figurename~\ref{fig-fcn_archi} is exactly identical to the architecture proposed in~\cite{wang2017time}. 
This enabled us to solely test the effect of transfer learning when fitting a deep learning model for the TSC task. 
We should also note that for fine-tuning and training from scratch, we are using the same network architecture with the same hyper-parameters, except for the last fully-connected layer whose adaptation is explained in the following subsection.

\subsection{Network adaptation}

After training the previously described network on the 85 datasets in the archive, we obtain 85 different neural networks.
The only difference between these 85 neural network architectures lies in the output layer.
The rest of the layers have the same number of parameters but with different values. 
In fact the last layer, which is a softmax classifier, depends on the number of classes in the dataset.

Thus, given a source dataset $D_s$ and a target dataset $D_t$, we first train the network on $D_s$. 
We then remove the last layer and replace it with another softmax layer whose number of neurons is equal to the number of classes in the target dataset $D_t$.  
The added softmax layer's parameters are initialized randomly using Glorot's uniform initialization~\cite{glorot2010understanding}.
This new network is then re-trained (fine-tuned) on $D_t$.

We chose to fine-tune the whole network instead of training only the last newly added output layer. 
We tried to limit back-propagating the gradient to the last layer, but found that the network failed to converge. 
This is in compliance with the transfer learning literature~\cite{yosinski2014transferable}, where re-training the whole network almost always leads to better results. 

Finally, we should add that one of the advantages of using a global average pooling layer is that we do not need to re-scale the input time series when transferring models between time series of different length. 

\begin{table}
\renewcommand{\arraystretch}{1.2}
	\centering
	\begin{tabular}{|c|c|c|}
    \hline
    \textbf{Hyperparameter} & \textbf{Original}~\cite{wang2017time} & \textbf{Ours} \\
    \toprule
    \bottomrule
    Epochs & 2000 & 2000\\ \hline
    Batch size & 16 & 16\\ \hline
    Optimizer & Adam~\cite{kingma2015adam} & Adam~\cite{kingma2015adam} \\ \hline
	Learning rate & 0.001 & 0.001 \\ \hline
    First moment & 0.9 & 0.9 \\ \hline
    Second moment & 0.999&0.999 \\ \hline
    Loss function  & Cross-entropy&Cross-entropy \\ \hline
    \end{tabular}
    \caption{Table showing the same hyperparameters for both approaches: with or without transfer learning.}
    \label{tab-hyper}
\end{table} 

\subsection{Inter-datasets similarity}

One of the main challenges with transfer learning is choosing the source dataset. 
In~\cite{pan2011transfer}, it was demonstrated that a learning algorithm trained with a certain source domain will not yield an optimal performance if the marginal distributions of the datasets' input are different.  
In our case, the total number of datasets in the UCR archive is 85. 
Therefore for each target dataset in the archive, we have 84 potential source datasets. 
This makes the trial and error based approach for transfer learning very costly in terms of computational resources.
Hence, we propose to use the DTW distance to compute the similarities between the datasets, thus guiding the choice of a source dataset for a given target dataset.

Note that it is practically impossible to directly estimate the performance of a model learned on a source dataset by applying it on a target dataset's train set since the last layer of the network is \emph{specific}~\cite{yosinski2014transferable} to the classes of the source dataset.

In order to compute the similarities between the datasets, we first reduce the number of time series for each dataset to one time series (or prototype) per class.
The per class prototype is computed by averaging the set of time series in the corresponding class. 
We used the well-known DTW Barycenter Averaging (DBA) method to the average a set of time series~\cite{petitjean2014summarizing}. 
The latter summarizing function was proposed and validated as an averaging method in the DTW induced space. 
In addition, DBA has been recently used as a data reduction technique where it was evaluated in a nearest centroid classification schema~\cite{petitjean2014dynamic}.
Therefore, to generate the similarity matrix between the UCR datasets, we computed a distance between each pair of datasets.
Finally, for simplicity and since the main goal of this paper is not the inter-datasets similarity,  we chose the distance between two datasets to be equal to the minimum distance between the prototypes of their corresponding classes. 

Algorithm~\ref{algo-sim} shows the different steps followed to compute the distance matrix between the UCR datasets.
The first part of the algorithm (lines 1 through 7) presents the data reduction technique similar to~\cite{petitjean2014dynamic}.
For the latter step, we first go through the classes of each dataset (lines 1, 2 and 3) and then average the set of time series for each class. 
Following the recommendations in~\cite{petitjean2014dynamic}, the averaging method (DBA) was initialized to be equal to the medoid of the time series selected set (line 4). 
We fixed the number of iterations for the DBA algorithm to be equal to 10, for which the averaging method has been shown to converge~\cite{petitjean2011a}.

After having reduced the different sets for each time series dataset, we proceed to the actual distance computation step (lines 8 through 22).
From line 8 to 10, we loop through every possible combination of datasets pairs. 
Lines 13 and 14 show the loop through each class for each dataset (at this stage each class is represented by one average time series thanks to the data reduction steps).
Finally, lines 15 through 19 set the distance between two datasets to be equal to the minimum DTW distance between their corresponding classes. 

One final note is that when computing the similarity between the datasets, the only time series data we used came from the training set, thus eliminating any bias due to having seen the test set's distribution. 

\begin{algorithm}
\begin{algorithmic}[1]
 \renewcommand{\algorithmicrequire}{\textbf{Input:}}
 \renewcommand{\algorithmicensure}{\textbf{Output:}}
 \REQUIRE $N$ time series datasets in an array $D$
 \ENSURE  $N\times N$ datasets similarity matrix
 \\ \textit{Initialization} : matrix $M$ of size $N\times N$ 
 \\ \textit{data reduction step}
  \FOR {$i = 1 $ to $N$}
  	\STATE $C=D[i].classes$
  	\FOR {$c=1$ to $ length(C)$}
    	\STATE $avg\_init = medoid(C[c])$
        \STATE $C[c] = DBA(C[c],avg\_init)$ 
	\ENDFOR
  \ENDFOR
  \\ \textit{distance calculation step}
  \FOR {$i = 1 $ to $N$}
  	\STATE $C_i=D[i].classes$
  	\FOR {$j=1$ to $N$}
  		\STATE $C_j=D[j].classes$
    	\STATE $dist = \infty$
        \FOR {$c_i=1$ to $ length(C_i)$}
        	\FOR{$c_j=1$ to $ length(C_j)$}
        		\STATE $cdist = DTW(C_i[c_i], C_j[c_j])$
        		\STATE $dist = minimum(dist, cdist)$ 
            \ENDFOR
        \ENDFOR
        \STATE $M[i,j] = dist$
    \ENDFOR
  \ENDFOR
 \RETURN $M$ 
\end{algorithmic}
\caption{Inter-datasets similarity}
\label{algo-sim}
\end{algorithm}

\section{Experimental setup}\label{sec-experiments}

\subsection{Datasets}

We evaluate our developed framework thoroughly on the largest publicly available benchmark for time series analysis: the UCR archive~\cite{ucrarchive}, which consists of 85 datasets selected from various real-world domains.
The time series in the archive are already z-normalized to have a mean equal to zero and a standard deviation equal to one. 
During the experiments, we used the default training and testing set splits provided by UCR. 
For pre-training a model, we used only the train set of the source dataset.
We also fine-tuned the pre-trained model solely on the target dataset's training data. 
Hence the test sets were only used for evaluation purposes.    



\subsection{Experiments}

For each pair of datasets ($D_1$ and $D_2$) in the UCR archive we need to perform two experiments:
\begin{itemize}
	\item $D_1$ is the source dataset and $D_2$ is the target dataset. 
    \item $D_1$ is the target dataset and $D_2$ is the source dataset. 
\end{itemize}
Which makes it in total 7140 experiments for the 85 dataset in the archive. 
Hence, given the huge number of models that need to be trained, we ran our experiments on a cluster of 60 GPUs. 
These GPUs were a mix of three types of Nvidia graphic cards: GTX 1080 Ti, Tesla K20, K40 and K80. 
The total sequential running time was approximately 168 days, that is if the computation has been done on a single GPU.  
But by leveraging the cluster of 60 GPUs, we managed to obtain the results in less than one week.
We implemented our framework using the open source deep learning library Keras~\cite{keras} with the Tensorflow~\cite{tensorflow} back-end.
For reproducibility purposes, we provide the 7140 trained Keras models (in a HDF5 format) on the companion web page of the paper\footnote{\url{http://germain-forestier.info/src/bigdata2018/}}. We have also published the raw results and the full source code of our method to enable the time series community to verify and build upon our findings\footnote{\url{https://github.com/hfawaz/bigdata18}}.

\begin{figure*}
\begin{center}
  \includegraphics[width=\textwidth]{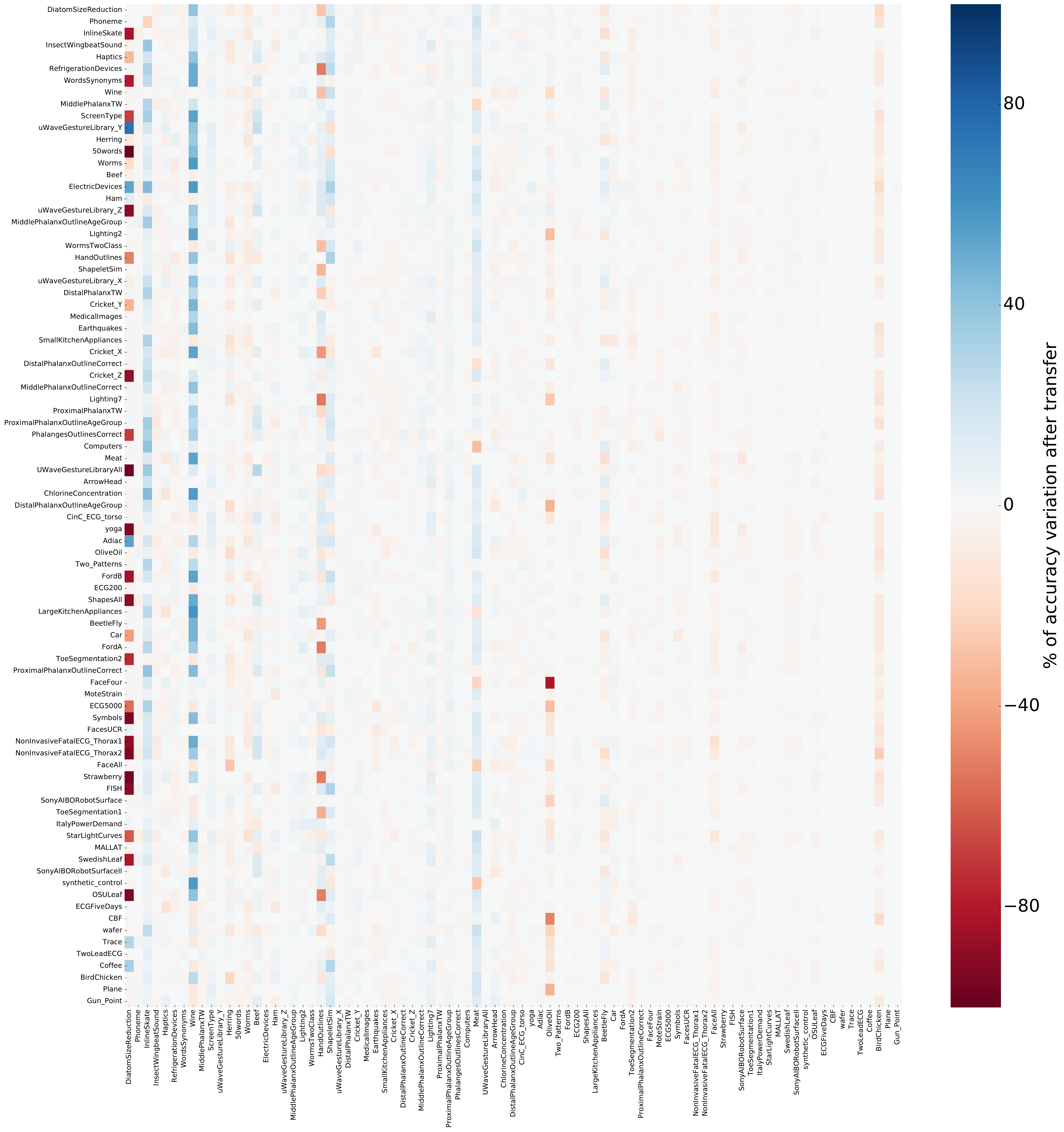}
  \caption{
  The variation in percentage over the original accuracy when fine tuning a pre-trained model.
  The rows' indexes correspond to the source datasets and the columns' indexes correspond to the target datasets.
  The \textbf{red} color shows the extreme case where the chosen pair of datasets (source and target) deteriorates the network's performance.
  Where on the other hand, the \textbf{blue} color identifies the improvement in accuracy when transferring the model from a certain source dataset and fine-tuning on another target dataset.
  The white color means that no change in accuracy has been identified when using the transfer learning method for two datasets. 
  The matrix actually has a size of $85\times85$ (instead of $85\times84$) for visual clarity with its diagonal left out of the analysis.
  (Best viewed in color).
  }
  \label{fig-heat_map}
\end{center}
\end{figure*}

\section{Results}\label{sec-results}

The experiments described in the previous section yielded interesting yet hard-to-understand results. 
In this section, we first present the result of the $85\times84$ experiments in a form of a matrix (displayed as a heat map in \figurename~\ref{fig-heat_map}). 
We then empirically show how choosing the wrong source dataset for a given target dataset could decrease the network's performance. 
Therefore, we provide a DTW based solution to choose the best source dataset for a given target dataset. 
Finally, we detail a few interesting case studies where the behavior of the proposed method has a significant impact on the transfered model's accuracy. 

\subsection{Transfer learning accuracy variation matrix}

In order to have a fair comparison across the datasets, we illustrate the variation in the transferred model's accuracy based on the percentage of variation compared to the original accuracy (without transfer learning). 
For example, consider the original accuracy (equal to 74.6\%) when training the neural network from scratch on the target dataset HandOutlines. 
Then instead of training the model from scratch (with random initializations) we obtain a 86.5\% accuracy when initializing the network's weights to be equal to the weights of a pre-trained network on the source dataset MedicalImages.
Hence, the percentage of accuracy variation with respect to the original value is equal to $100\times(86.5-74.6)/74.6\approx+16\%$. 
Thus negative values (red in \figurename~\ref{fig-heat_map}) indicate a decrease in performance when using the transfer learning approach. 
Whereas, a positive percentage (blue in \figurename~\ref{fig-heat_map}) indicates an increase in performance when fine-tuning a pre-trained model. 

When observing the heat map in \figurename~\ref{fig-heat_map}, one can easily see that fine-tuning a pre-trained model almost never hurts the performance of the CNN.
This can be seen by the dominance of the white color in the heat map, which corresponds to almost no variation in accuracy. 

On the other hand, the results which we found interesting are the two extreme cases (red and blue) where the use of transfer learning led to high variations in accuracy.
Interestingly for a given target dataset, the choice of source dataset could deteriorate or improve the CNN's performance as we will see in the following subsection. 

\subsection{Naive transfer learning}

\begin{figure}[htbp]
\centerline{\includegraphics[width=0.95\linewidth]{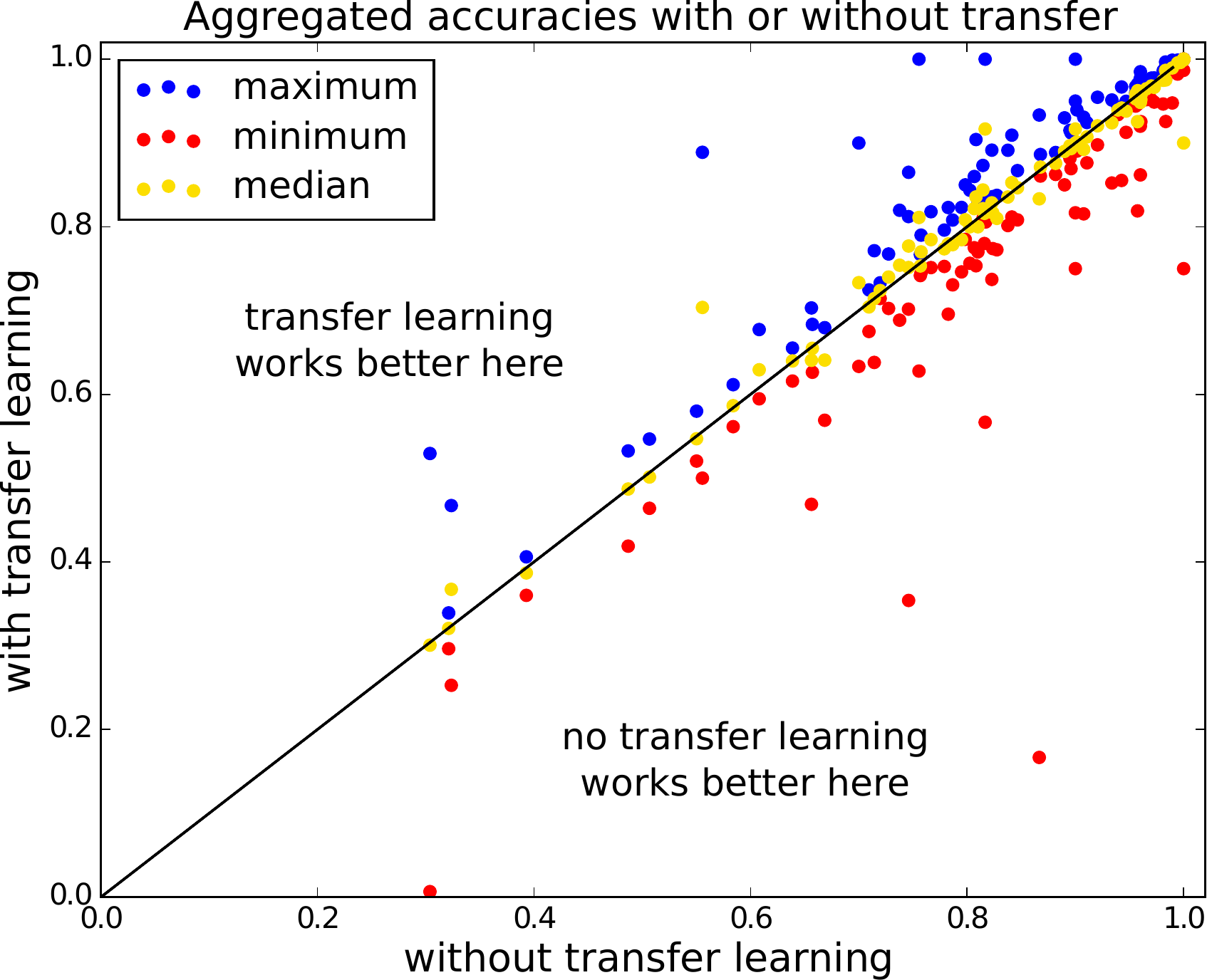}}
\caption{The three aggregated accuracies (minimum, median and maximum) of the Convolutional Neural Networks with the transfer learning approach against no transfer learning.}
\label{fig-min_max}
\end{figure}

While observing the heat map in \figurename~\ref{fig-heat_map}, we can easily see that certain target datasets (columns) exhibit a high variance of accuracy improvements when varying the source datasets.
Therefore, to visualize the worst and best case scenarios when fine-tuning a model against training from scratch, we plotted in \figurename~\ref{fig-min_max} a pairwise comparison of three aggregated accuracies \{$minimum,median,maximum$\}. 

For each target dataset $D_t$, we took its minimum accuracy among the source datasets and plot it against the model's accuracy when trained from scratch.
This corresponds to the red dots in \figurename~\ref{fig-min_max}.  
By taking the minimum, we illustrate how one can \emph{always} find a bad source dataset for a given target dataset and decrease the model's original accuracy when fine-tuning a pre-trained network.
  
On the other hand, the maximum accuracy (blue dots in \figurename~\ref{fig-min_max}) shows that there is also \emph{always} a case where a source dataset increases the accuracy when using the transfer learning approach. 

As for the median (yellow dots in \figurename~\ref{fig-min_max}), it shows that on average, pre-training and then fine-tuning a model on a target dataset improves without significantly hurting the model's performance. 

One extreme case, where the choice of the source dataset had a huge impact on the model's accuracy, is the OliveOil dataset.
Precisely the accuracy decreased from 93.3\% to 16.7\% when choosing respectively MALLAT and FaceFour as source datasets. 

This analysis showed us that blindly and naively using the transfer learning approach could drastically decrease the model's performance. 
Actually, this is largely due to the fact that the initial weights of the network have a significant impact on the training~\cite{glorot2010understanding}. 
This problem has been identified as \emph{negative transfer learning} in the literature, where there still exists a need to quantify the amount of relatedness between the source and target datasets and whether an attempt to transfer knowledge from the source to the target domain should be made~\cite{weiss2016a}. 
Therefore in the following subsection, we show how our similarity based solution can quantify this relatedness between the source and the target, thus enabling us to predict the best source dataset for a given target dataset. 

\subsection{Smart transfer learning}

In order to know in advance which source dataset is suited for which target dataset, we propose to leverage the similarity between two datasets. 
Our method is designed specifically for time series data without any previous domain knowledge about the datasets. 
Using the method we described in Section~\ref{sec-method}, we managed to compute a nearest neighbor for a target dataset and set this nearest neighbor to be the chosen source dataset for the current target dataset in question. 

The results showed that this proposed DTW based method will help in achieving what is called \emph{positive transfer}~\cite{weiss2016a}. 
As opposed to \emph{negative transfer}, positive transfer learning means that the learning algorithm's accuracy increases when fine-tuning a pre-trained model compared to a training from scratch approach~\cite{weiss2016a}.

\begin{figure}[htbp]
\centerline{\includegraphics[width=0.95\linewidth]{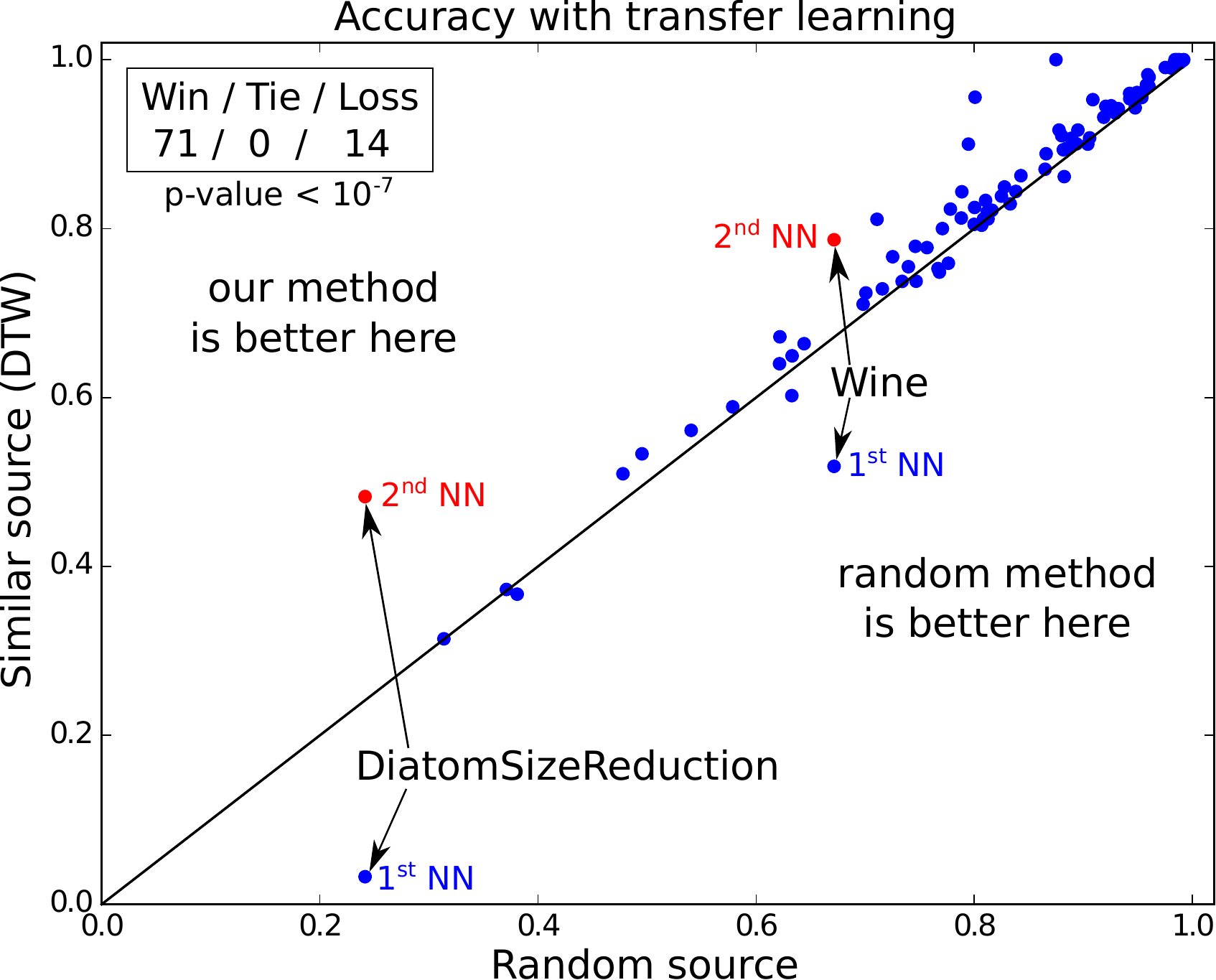}}
\caption{
The accuracy of a fine-tuned model for two cases: 
($x$ axis) when the source dataset is selected randomly;  
($y$ axis) when the source dataset is selected using our Dynamic Time Warping based solution.}
\label{fig-dtw_vs_rnd}
\end{figure}

\figurename~\ref{fig-dtw_vs_rnd} shows a pairwise accuracy plot for two approaches: a random selection process of the source dataset against a ``smart'' selection of the source dataset using a nearest neighbor algorithm with the distance calculated in algorithm~\ref{algo-sim}. 
In order to reduce the bias due to the random seed, the accuracy for the random selection approach was averaged over 1000 iterations. 
This plot shows that on average, choosing the most similar dataset using our method is significantly better than a random selection approach (with $p<10^{-7}$ for the Wilcoxon signed-rank test).
Respectively our method wins, ties and loses on 71, 0 and 14 datasets against randomly choosing the source dataset.
We should also note that for the two datasets DiatomSizeReduction and Wine, the nearest neighbor is not always the best choice.  
Actually, we found that the second nearest neighbor increases drastically the accuracy from 3.3\% to 46.7\% for DiatomSizeReduction and from 51.9\% to 77.8\% for Wine (see the $2^{nd}$ NN dots in \figurename~\ref{fig-dtw_vs_rnd}). 
This means that certain improvements could be incorporated to our inter-datasets similarity calculation such as adding a warping window~\cite{dau2017judicious} or changing the number of prototypes for each class which we aim to study in our future work.

Therefore, since in \figurename~\ref{fig-dtw_vs_rnd} the most similar dataset is the only one that is considered as a potential source for a given target, another interesting study would be to analyze the accuracy on a given target dataset as a function of how dissimilar the source dataset is.
However due to the huge number of datasets in the UCR archive compared to the space limitation, we chose to only study the most interesting cases where the results can be visually interpreted.
The analysis for the whole 85 datasets is however included in the companion web page.

\subsection{Interesting case studies}

In this final analysis we chose to work with three interesting target datasets: \textit{ShapeletSim}, \textit{HandOutlines} and \textit{Meat}.
These datasets were chosen for different reasons such as the small size of the training set, the relatedness to shapelets and the transfer learning's accuracy variation.

\begin{figure}[htbp]
\centerline{\includegraphics[width=0.95\linewidth]{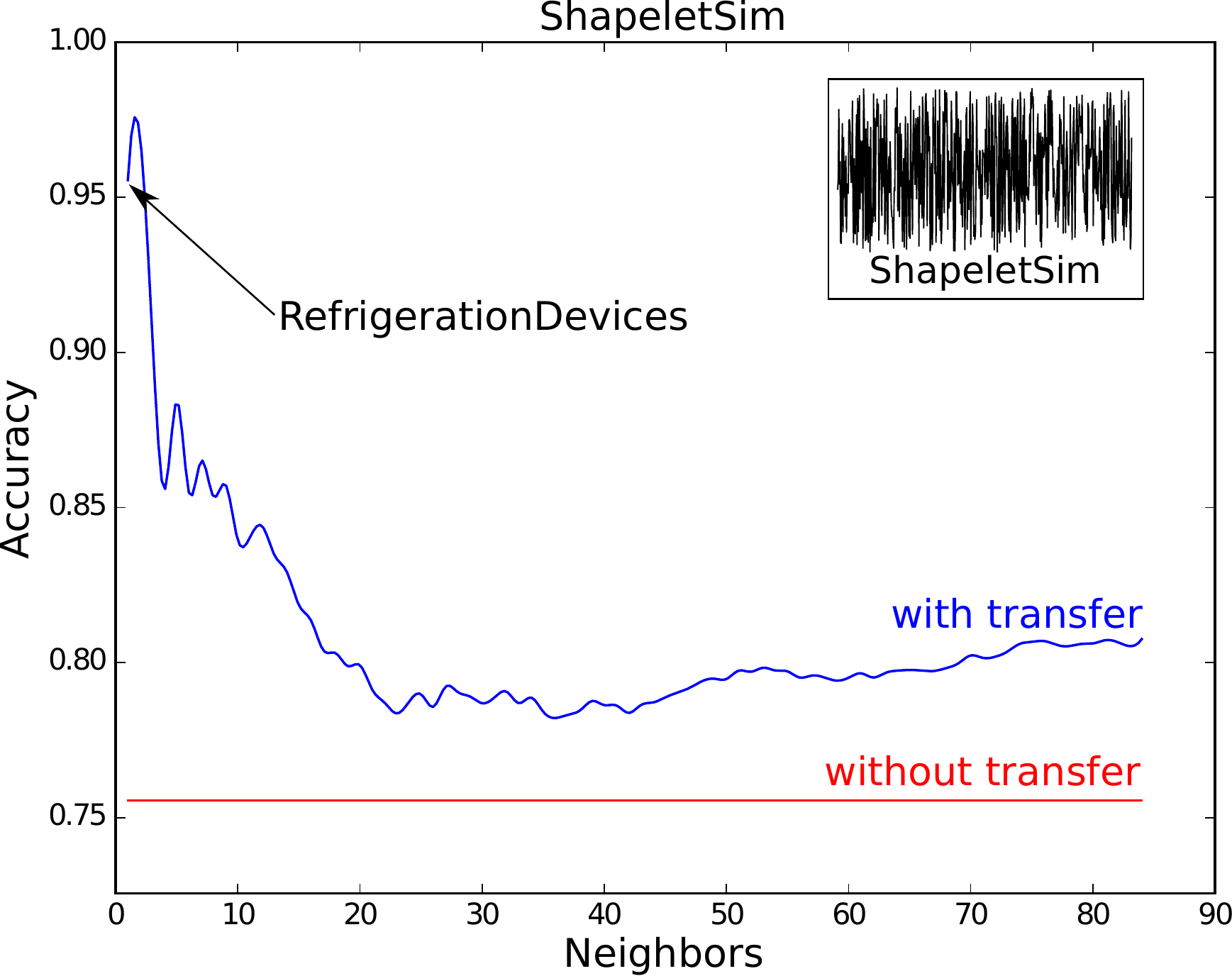}}
\caption{
The fine-tuned model's accuracy variation on the target dataset ShapeletSim with respect to the chosen source dataset neighbor (smoothed for visual clarity - best viewed in color).
}
\label{fig-ShapeletSim}
\end{figure}

\textit{ShapeletSim} contains one of the smallest training sets in the UCR archive (with 20 training instances). 
Additionally, this dataset is a simulated dataset designed specifically for shapelets which makes it interesting to see how well CNNs can fine-tune (pre-learned) shapelets~\cite{cui2016multi} when varying the source dataset.
\figurename~\ref{fig-ShapeletSim} shows how the model's accuracy decreases as we go further from the target dataset.
Precisely the average accuracy for the top 3 neighbors reaches 93\% compared to the original accuracy of 76\%. 
Actually, we found that the closest dataset to \textit{ShapeletSim} is the RefrigerationDevices dataset which contains readings from 251 households with the task to identify three classes: Fridge, Refrigerator and Upright Freezer.
This is very interesting since using other background knowledge one cannot easily predict that using RefrigerationDevices as a source for ShapeletSim will lead to better accuracy improvement. 
To understand better this source/target association, we investigated the shapes of the time series of each dataset and found that both datasets exhibit very similar spiky subsequences which is likely the cause for the transfer learning to work between these two datasets. 

\begin{figure}[htbp]
\centerline{\includegraphics[width=0.95\linewidth]{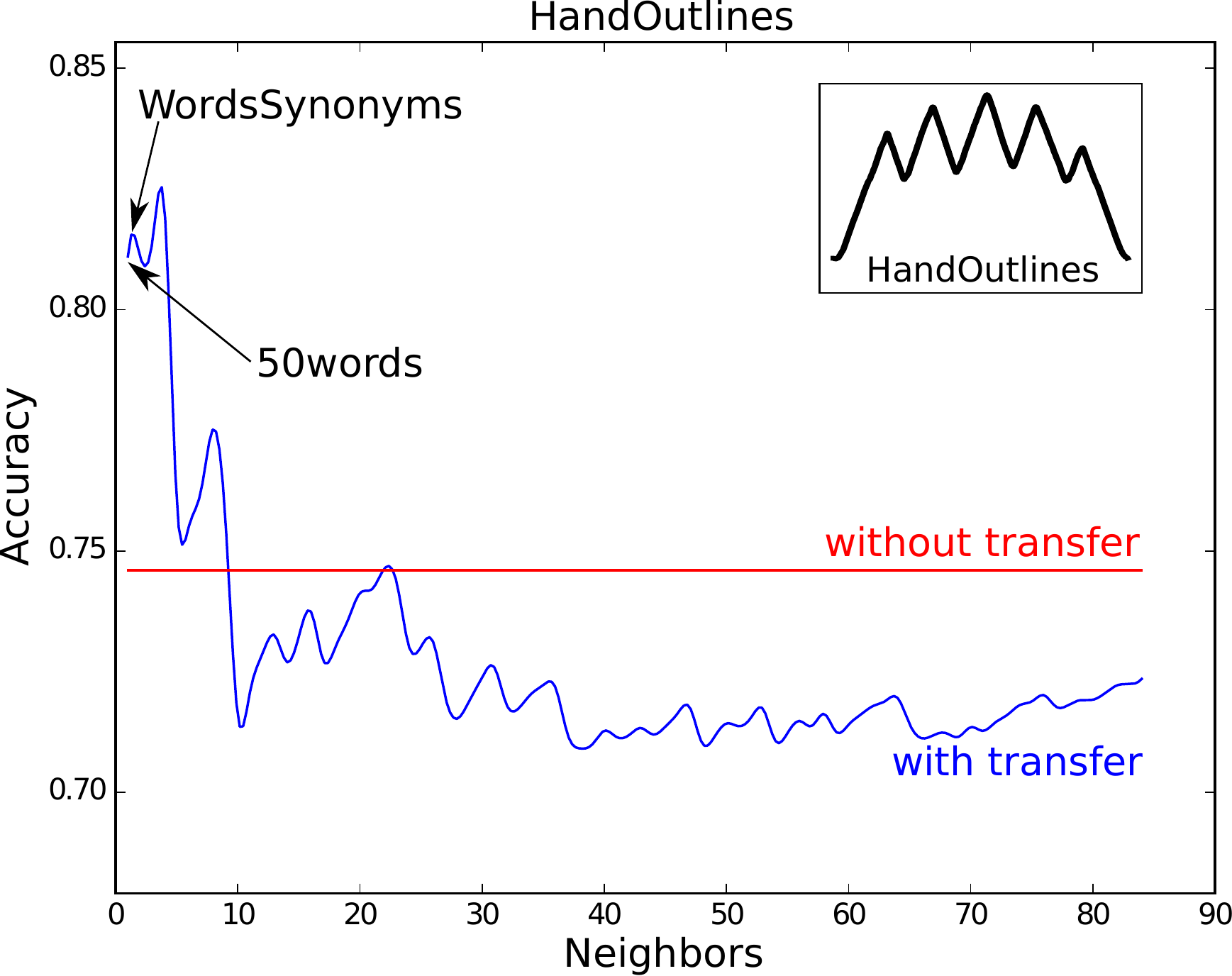}}
\caption{
The fine-tuned model's accuracy variation on the target dataset HandOutlines with respect to the chosen source dataset neighbor (smoothed for visual clarity - best viewed in color).
}
\label{fig-HandOutlines}
\end{figure}

\textit{HandOutlines} is one of the datasets where fine-tuning a pre-trained model almost never improves the accuracy.
Unlike \textit{ShapeletSim}, this dataset contains enough labeled data for the learning algorithm to learn from (with 1000 time series in the training set).
Surprisingly, we found that one could drastically increase the model's performance when choosing the best source dataset. 
\figurename~\ref{fig-HandOutlines} shows a huge difference (10\%) between the model's accuracy when fine-tuned using the most \emph{similar} source dataset and the accuracy when choosing the most \emph{dissimilar} source dataset.
\textit{HandOutlines} is a classification problem that uses the outlines extracted from hand images. 
We found that the two most similar datasets (50words and WordsSynonyms) that yielded high accuracy improvements, are also words' outlines extracted from images of George Washington's manuscripts. 

\begin{figure}[htbp]
\centerline{\includegraphics[width=0.95\linewidth]{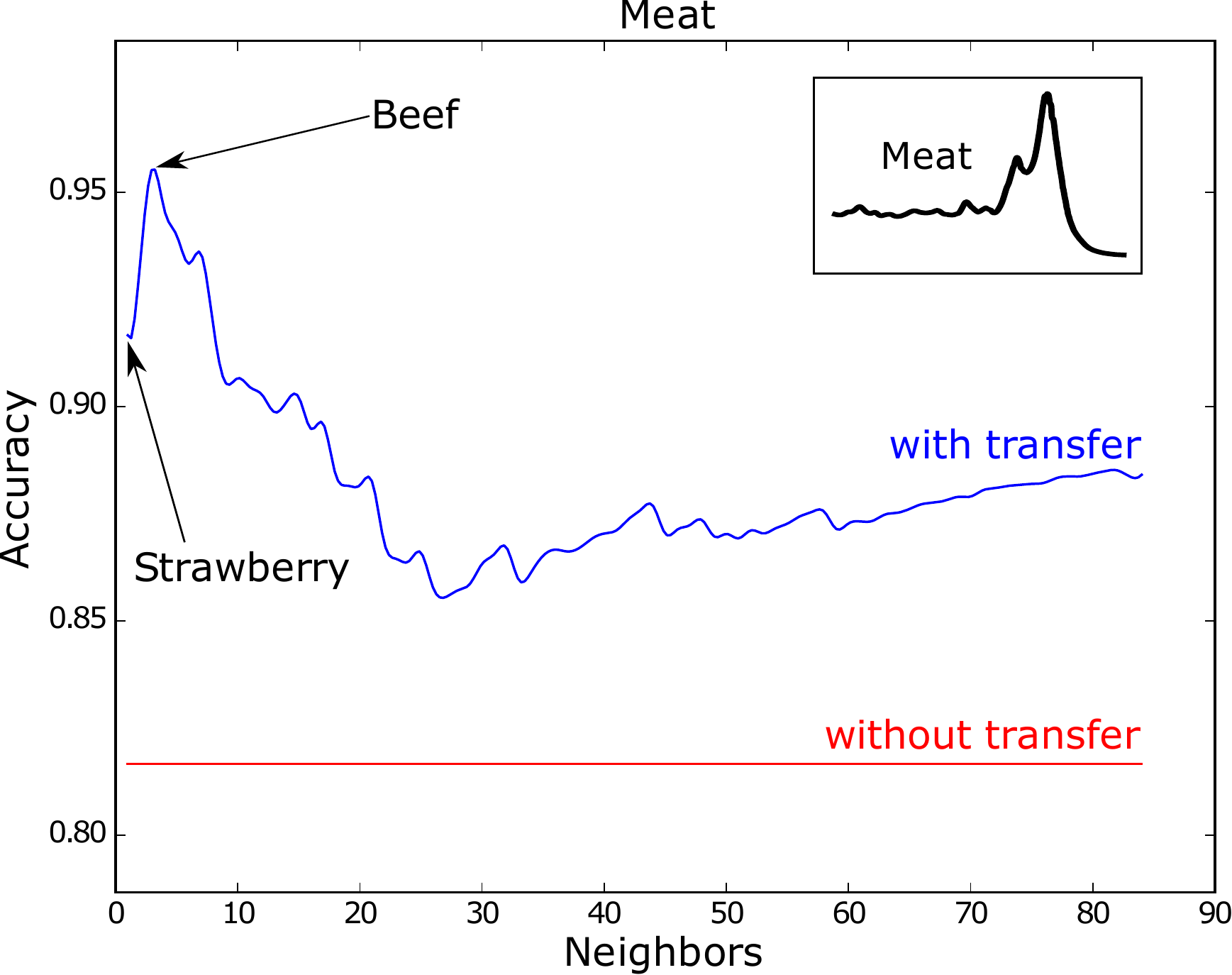}}
\caption{
The fine-tuned model's accuracy variation on the target dataset Meat with respect to the chosen source dataset neighbor (smoothed for visual clarity - best viewed in color).
}
\label{fig-Meat}
\end{figure}

\textit{Meat} is one of the smallest datasets (with 20 training instances) where the transfer learning approach was almost always beneficial.
However, we would like to examine the possibility of improving the accuracy even for the case where the transfer learning seems to be positive~\cite{weiss2016a} for any choice of source dataset.
\figurename~\ref{fig-Meat} shows that the accuracy reaches almost 95\% for the top 3 closest datasets and then decrease the less similar the source and target datasets are.
While investigating these similarities, we found the top 1 and 3 datasets to be respectively Strawberry and Beef which are all considered spectrograph datasets~\cite{bagnall2017the}. 
As for the second most similar dataset, our method determined it was 50words.
Given the huge number of classes (fifty) in 50words our method managed to find some latent similarity between the two datasets which helped in improving the accuracy of the transfer learning process.

\section{Conclusion}\label{sec-conclusion}

In this paper, we investigated the transfer learning approach on a state of the art deep learning model for TSC problems. 
Our extensive experiments with every possible combination of source and target datasets in the UCR archive, were evidence that the choice of the source dataset could have a significant impact on the model's generalization capabilities. 
Precisely when choosing a bad source dataset for a given target dataset, the optimization algorithm can be stuck in a local optimum. 
This phenomena has been identified in the transfer learning literature by \emph{negative transfer learning} which is still an active area of research~\cite{weiss2016a}.
Thus, when deploying a transfer learning approach, the big data practitioner should give attention to the relationship between the target and the chosen source domains.

These observations motivated us to examine the use of the well known time series similarity measure DTW, to predict the choice of the source dataset when fine-tuning a model on a time series target dataset.
After applying this transfer learning guidance, we concluded that transferring deep CNNs on a target dataset works best when fine-tuning a network that was pre-trained on a similar source dataset.
These findings are very interesting since no previous observation made the link between the space induced by the classic DTW and the features learned by the Convolutional Neural Networks. 

Our results should motivate the big data practitioners to no longer train models from scratch when classifying time series, but instead to fine-tune pre-trained models. 
Especially because CNNs, if designed properly, can be adapted across different time series datasets with varying length.  

In our future work, we aim again to reduce the deep neural network's overfitting phenomena by generating synthetic data using a Weighted DTW Barycenter Averaging method~\cite{forestier2017generating}, since the latter distance gave encouraging results in guiding a complex deep learning tool such as transfer learning.    

Finally, with big data repositories becoming more frequent, leveraging existing source datasets that are similar to, but not exactly the same as a target dataset of interest, makes a transfer learning method an enticing approach.   

\section*{Acknowledgments}
The authors would like to thank NVIDIA Corporation for the GPU Grant and the M\'esocentre of Strasbourg for providing access to the GPU cluster. 


\bibliographystyle{IEEEtran}
\bibliography{biblio}

\begin{thebibliography}{10}
\providecommand{\url}[1]{#1}
\csname url@samestyle\endcsname
\providecommand{\newblock}{\relax}
\providecommand{\bibinfo}[2]{#2}
\providecommand{\BIBentrySTDinterwordspacing}{\spaceskip=0pt\relax}
\providecommand{\BIBentryALTinterwordstretchfactor}{4}
\providecommand{\BIBentryALTinterwordspacing}{\spaceskip=\fontdimen2\font plus
\BIBentryALTinterwordstretchfactor\fontdimen3\font minus
  \fontdimen4\font\relax}
\providecommand{\BIBforeignlanguage}[2]{{%
\expandafter\ifx\csname l@#1\endcsname\relax
\typeout{** WARNING: IEEEtran.bst: No hyphenation pattern has been}%
\typeout{** loaded for the language `#1'. Using the pattern for}%
\typeout{** the default language instead.}%
\else
\language=\csname l@#1\endcsname
\fi
#2}}
\providecommand{\BIBdecl}{\relax}
\BIBdecl

\bibitem{ucrarchive}
Y.~Chen, E.~Keogh, B.~Hu, N.~Begum, A.~Bagnall, A.~Mueen, and G.~Batista,
  ``{The UCR Time Series Classification Archive},'' July 2015.

\bibitem{wang2017time}
Z.~Wang, W.~Yan, and T.~Oates, ``Time series classification from scratch with
  deep neural networks: A strong baseline,'' in \emph{International Joint
  Conference on Neural Networks}, 2017, pp. 1578--1585.

\bibitem{bagnal2015time}
A.~Bagnall, J.~Lines, J.~Hills, and A.~Bostrom, ``{Time-Series Classification
  with {COTE}: The Collective of Transformation-Based Ensembles},'' \emph{IEEE
  Transactions on Knowledge and Data Engineering}, vol.~27, no.~9, pp.
  2522--2535, 2015.

\bibitem{zhang2016understanding}
C.~{Zhang}, S.~{Bengio}, M.~{Hardt}, B.~{Recht}, and O.~{Vinyals},
  ``{Understanding deep learning requires rethinking generalization},''
  \emph{ArXiv}, 2016.

\bibitem{IsmailFawaz2018data}
H.~Ismail~Fawaz, G.~Forestier, J.~Weber, L.~Idoumghar, and P.-A. Muller, ``Data
  augmentation using synthetic data for time series classification with deep
  residual networks,'' in \emph{International Workshop on Advanced Analytics
  and Learning on Temporal Data, {ECML} {PKDD}}, 2018.

\bibitem{yosinski2014transferable}
J.~Yosinski, J.~Clune, Y.~Bengio, and H.~Lipson, ``How transferable are
  features in deep neural networks?'' in \emph{Advances in neural information
  processing systems}, 2014, pp. 3320--3328.

\bibitem{csurka2017domain}
G.~{Csurka}, ``{Domain Adaptation for Visual Applications: A Comprehensive
  Survey},'' \emph{ArXiv e-prints}, 2017.

\bibitem{russakovsky2015imagenet}
O.~Russakovsky, J.~Deng, H.~Su, J.~Krause, S.~Satheesh, S.~Ma, Z.~Huang,
  A.~Karpathy, A.~Khosla, M.~Bernstein, A.~C. Berg, and L.~Fei-Fei, ``{ImageNet
  Large Scale Visual Recognition Challenge},'' \emph{International Journal of
  Computer Vision}, vol. 115, no.~3, pp. 211--252, 2015.

\bibitem{openimages}
I.~Krasin, T.~Duerig, N.~Alldrin, V.~Ferrari, S.~Abu-El-Haija, A.~Kuznetsova,
  H.~Rom, J.~Uijlings, S.~Popov, S.~Kamali, M.~Malloci, J.~Pont-Tuset, A.~Veit,
  S.~Belongie, V.~Gomes, A.~Gupta, C.~Sun, G.~Chechik, D.~Cai, Z.~Feng,
  D.~Narayanan, and K.~Murphy, ``{OpenImages}: A public dataset for large-scale
  multi-label and multi-class image classification.'' \emph{Dataset available
  from https://storage.googleapis.com/openimages/web/index.html}, 2017.

\bibitem{cui2016multi}
Z.~{Cui}, W.~{Chen}, and Y.~{Chen}, ``{Multi-Scale Convolutional Neural
  Networks for Time Series Classification},'' \emph{ArXiv}, 2016.

\bibitem{gamboa2017deep}
J.~{Cristian Borges Gamboa}, ``{Deep Learning for Time-Series Analysis},''
  \emph{ArXiv}, 2017.

\bibitem{ye2009time}
L.~Ye and E.~Keogh, ``{Time Series Shapelets: A New Primitive for Data
  Mining},'' in \emph{ACM SIGKDD International Conference on Knowledge
  Discovery and Data Mining}, 2009, pp. 947--956.

\bibitem{grabocka2014learning}
J.~Grabocka, N.~Schilling, M.~Wistuba, and L.~Schmidt-Thieme, ``{Learning
  Time-series Shapelets},'' in \emph{ACM SIGKDD International Conference on
  Knowledge Discovery and Data Mining}, 2014, pp. 392--401.

\bibitem{weiss2016a}
K.~Weiss, T.~M. Khoshgoftaar, and D.~Wang, ``A survey of transfer learning,''
  \emph{Journal of Big Data}, vol.~3, no.~1, p.~9, 2016.

\bibitem{krizhevsky2012imagenet}
A.~Krizhevsky, I.~Sutskever, and G.~E. Hinton, ``{ImageNet Classification with
  Deep Convolutional Neural Networks},'' in \emph{Advances in Neural
  Information Processing Systems 25}, 2012, pp. 1097--1105.

\bibitem{lecun2015deep}
Y.~LeCun, Y.~Bengio, and G.~Hinton, ``Deep learning,'' \emph{Nature}, vol. 521,
  pp. 436--444, 2015.

\bibitem{szegedy2015going}
C.~Szegedy, W.~Liu, Y.~Jia, P.~Sermanet, S.~Reed, D.~Anguelov, D.~Erhan,
  V.~Vanhoucke, and A.~Rabinovich, ``Going deeper with convolutions,'' in
  \emph{IEEE Conference on Computer Vision and Pattern Recognition}, 2015, pp.
  1--9.

\bibitem{sutskever2014sequence}
I.~Sutskever, O.~Vinyals, and Q.~V. Le, ``{Sequence to Sequence Learning with
  Neural Networks},'' in \emph{Neural Information Processing Systems}, 2014,
  pp. 3104--3112.

\bibitem{bahdanau2014neural}
D.~{Bahdanau}, K.~{Cho}, and Y.~{Bengio}, ``{Neural Machine Translation by
  Jointly Learning to Align and Translate}.''

\bibitem{geng2018cost}
Y.~{Geng} and X.~{Luo}, ``{Cost-Sensitive Convolution based Neural Networks for
  Imbalanced Time-Series Classification},'' \emph{ArXiv e-prints}, 2018.

\bibitem{leguennec2016data}
A.~Le~Guennec, S.~Malinowski, and R.~Tavenard, ``{Data Augmentation for Time
  Series Classification using Convolutional Neural Networks},'' in
  \emph{{ECML/PKDD Workshop on Advanced Analytics and Learning on Temporal
  Data}}, 2016.

\bibitem{langkvist2014a}
M.~L\"{a}ngkvist, L.~Karlsson, and A.~Loutfi, ``A review of unsupervised
  feature learning and deep learning for time-series modeling,'' \emph{Pattern
  Recognition Letters}, vol.~42, pp. 11 -- 24, 2014.

\bibitem{ziat2017spatio}
A.~Ziat, E.~Delasalles, L.~Denoyer, and P.~Gallinari, ``{Spatio-Temporal Neural
  Networks for Space-Time Series Forecasting and Relations Discovery},'' in
  \emph{IEEE International Conference on Data Mining}, 2017, pp. 705--714.

\bibitem{nweke2018deep}
H.~F. Nweke, Y.~W. Teh, M.~A. Al-garadi, and U.~R. Alo, ``Deep learning
  algorithms for human activity recognition using mobile and wearable sensor
  networks: State of the art and research challenges,'' \emph{Expert Systems
  with Applications}, vol. 105, pp. 233 -- 261, 2018.

\bibitem{che2017boosting}
Z.~Che, Y.~Cheng, S.~Zhai, Z.~Sun, and Y.~Liu, ``{Boosting Deep Learning Risk
  Prediction with Generative Adversarial Networks for Electronic Health
  Records},'' in \emph{IEEE International Conference on Data Mining}, 2017, pp.
  787--792.

\bibitem{IsmailFawaz2018deep}
H.~Ismail~Fawaz, G.~Forestier, J.~Weber, L.~Idoumghar, and P.-A. Muller, ``Deep
  learning for time series classification: a review,'' \emph{ArXiv}, 2018.

\bibitem{pan2010a}
S.~J. Pan and Q.~Yang, ``{A Survey on Transfer Learning},'' \emph{IEEE
  Transactions on Knowledge and Data Engineering}, vol.~22, no.~10, pp.
  1345--1359, 2010.

\bibitem{long2015learning}
M.~Long, Y.~Cao, J.~Wang, and M.~Jordan, ``{Learning Transferable Features with
  Deep Adaptation Networks},'' in \emph{International Conference on Machine
  Learning}, vol.~37, 2015, pp. 97--105.

\bibitem{ariefang2017da}
I.~B. Arief-Ang, F.~D. Salim, and M.~Hamilton, ``{DA-HOC: Semi-supervised
  Domain Adaptation for Room Occupancy Prediction Using CO2 Sensor Data},'' in
  \emph{International Conference on Systems for Energy-Efficient Built
  Environments}, 2017, pp. 1--10.

\bibitem{kasteren08recognizing}
T.~L. M.~V. Kasteren, G.~Englebienne, and B.~J.~A. Kröse, ``Recognizing
  activities in multiple contexts using transfer learning,'' in
  \emph{Association for the Advancement of Artificial Intelligence - AI in
  Elder care}, 2008, pp. 142--149.

\bibitem{vercruyssen2017transfer}
V.~Vercruyssen, W.~Meert, and J.~Davis, ``{Transfer Learning for Time Series
  Anomaly Detection},'' in \emph{Workshop and Tutorial on Interactive Adaptive
  Learning co-located with European Conference on Machine Learning and
  Principles and Practice of Knowledge Discovery in Databases}, 2017, pp.
  27--36.

\bibitem{spiegel2016}
S.~Spiegel, ``{Transfer Learning for Time Series Classification in
  Dissimilarity Spaces},'' in \emph{European Conference on Machine Learning and
  Principles and Practice of Knowledge Discovery in Databases}, 2016.

\bibitem{hu2016transfer}
Q.~Hu, R.~Zhang, and Y.~Zhou, ``Transfer learning for short-term wind speed
  prediction with deep neural networks,'' \emph{Renewable Energy}, vol.~85, pp.
  83 -- 95, 2016.

\bibitem{banerjee2017a}
D.~Banerjee, K.~Islam, G.~Mei, L.~Xiao, G.~Zhang, R.~Xu, S.~Ji, and J.~Li, ``{A
  Deep Transfer Learning Approach for Improved Post-Traumatic Stress Disorder
  Diagnosis},'' in \emph{IEEE International Conference on Data Mining}, 2017,
  pp. 11--20.

\bibitem{serra2018towards}
J.~{Serr{\`a}}, S.~{Pascual}, and A.~{Karatzoglou}, ``{Towards a universal
  neural network encoder for time series},'' \emph{ArXiv}, 2018.

\bibitem{bagnall2017the}
A.~Bagnall, J.~Lines, A.~Bostrom, J.~Large, and E.~Keogh, ``The great time
  series classification bake off: a review and experimental evaluation of
  recent algorithmic advances,'' \emph{Data Mining and Knowledge Discovery},
  vol.~31, no.~3, pp. 606--660, 2017.

\bibitem{ioffe2015batch}
S.~Ioffe and C.~Szegedy, ``{Batch Normalization: Accelerating Deep Network
  Training by Reducing Internal Covariate Shift},'' in \emph{International
  Conference on Machine Learning}, 2015, pp. 448--456.

\bibitem{zhou2016learning}
B.~Zhou, A.~Khosla, A.~Lapedriza, A.~Oliva, and A.~Torralba, ``{Learning Deep
  Features for Discriminative Localization},'' in \emph{IEEE Conference on
  Computer Vision and Pattern Recognition}, 2016, pp. 2921--2929.

\bibitem{glorot2010understanding}
X.~Glorot and Y.~Bengio, ``Understanding the difficulty of training deep
  feedforward neural networks,'' in \emph{International Conference on
  Artificial Intelligence and Statistics}, vol.~9, 2010, pp. 249--256.

\bibitem{kingma2015adam}
D.~P. Kingma and J.~{Ba}, ``Adam: A method for stochastic optimization,'' in
  \emph{International Conference on Learning Representations}, 2015.

\bibitem{pan2011transfer}
W.~Pan, N.~N. Liu, E.~W. Xiang, and Q.~Yang, ``{Transfer Learning to Predict
  Missing Ratings via Heterogeneous User Feedbacks},'' in \emph{International
  Joint Conference on Artificial Intelligence}, 2011, pp. 2318--2323.

\bibitem{petitjean2014summarizing}
F.~Petitjean and P.~Gan{\c c}arski, ``Summarizing a set of time series by
  averaging: From steiner sequence to compact multiple alignment,''
  \emph{Theoretical Computer Science}, vol. 414, no.~1, pp. 76 -- 91, 2012.

\bibitem{petitjean2014dynamic}
F.~Petitjean, G.~Forestier, G.~I. Webb, A.~E. Nicholson, Y.~Chen, and E.~Keogh,
  ``{Dynamic Time Warping Averaging of Time Series Allows Faster and More
  Accurate Classification},'' in \emph{IEEE International Conference on Data
  Mining}, 2014, pp. 470--479.

\bibitem{petitjean2011a}
F.~Petitjean, A.~Ketterlin, and P.~Gançarski, ``A global averaging method for
  dynamic time warping, with applications to clustering,'' \emph{Pattern
  Recognition}, vol.~44, no.~3, pp. 678 -- 693, 2011.

\bibitem{keras}
F.~Chollet \emph{et~al.}, ``Keras,'' \url{https://keras.io}, 2015.

\bibitem{tensorflow}
M.~Abadi, P.~Barham, J.~Chen, Z.~Chen, A.~Davis, J.~Dean, M.~Devin,
  S.~Ghemawat, G.~Irving, M.~Isard, M.~Kudlur, J.~Levenberg, R.~Monga,
  S.~Moore, D.~G. Murray, B.~Steiner, P.~Tucker, V.~Vasudevan, P.~Warden,
  M.~Wicke, Y.~Yu, and X.~Zheng, ``{TensorFlow: A System for Large-scale
  Machine Learning},'' in \emph{USENIX Conference on Operating Systems Design
  and Implementation}, 2016, pp. 265--283.

\bibitem{dau2017judicious}
H.~A. Dau, D.~F. Silva, F.~Petitjean, G.~Forestier, A.~Bagnall, and E.~Keogh,
  ``Judicious setting of dynamic time warping's window width allows more
  accurate classification of time series,'' in \emph{IEEE International
  Conference on Big Data}, 2017, pp. 917--922.

\bibitem{forestier2017generating}
G.~Forestier, F.~Petitjean, H.~A. Dau, G.~I. Webb, and E.~Keogh, ``{Generating
  Synthetic Time Series to Augment Sparse Datasets},'' in \emph{IEEE
  International Conference on Data Mining}, 2017, pp. 865--870.

\end{thebibliography}

\end{document}